\documentclass[11pt,a4paper]{article}
\usepackage[hyperref]{emnlp-ijcnlp-2019}
\usepackage{times}
\usepackage{latexsym}

\usepackage[utf8]{inputenc} 
\usepackage[T1]{fontenc}    
\usepackage{hyperref}       
\usepackage{url}            
\usepackage{booktabs}       
\usepackage{amsfonts}       
\usepackage{nicefrac}       
\usepackage{microtype}      

\usepackage{times}

\usepackage{eqnarray,amsbsy,amsmath,amssymb,amsthm}
\usepackage{algorithm}
\usepackage{graphicx}

\usepackage{algorithmic}
\usepackage{multirow}

\usepackage{tikz}
\usepackage{csquotes}
\usetikzlibrary{bayesnet} 
\usepackage{bbm}

\usepackage{csquotes}
\usepackage{mathtools}
\usepackage{comment}
\usepackage{natbib}
\usepackage{enumitem} 

\graphicspath{{pictures/}}

\newcommand{\bx}{\mathbf{x}}

\newcommand{\by}{\mathbf{y}}

\newcommand{\balpha}{\boldsymbol{\alpha}}

\newcommand{\bmu}{\boldsymbol{\mu}}

\newcommand{\btheta}{\boldsymbol{\theta}}

\newcommand{\ie}{{\it i.e.\!}}

\newcommand{\old}[1]{}

\newcommand{\vtheta}{\bm{\theta}}
\newcommand{\vbeta}{\bm{\beta}}

\usepackage{soul}
\usepackage{color}
\definecolor{aquamarine}{rgb}{0.5, 1.0, 0.83}
\definecolor{Mycolor2}{HTML}{00F9DE}

\usepackage{bm}
\usepackage{multirow}
\usepackage{tikz}
\usetikzlibrary{positioning,chains,fit,shapes,calc}
\usepackage{pgfplots}

\definecolor{turquoise}{cmyk}{.43,0,.24,0} 
\definecolor{myblue}{RGB}{80,80,160}
\definecolor{mygreen}{RGB}{143,255,204}

\tikzstyle{plate caption} = [caption, node distance=0, inner sep=-0.pt,below left=1pt and 0pt of #1.south east]
\tikzstyle{plate} = [draw, rectangle, fit=#1]

\graphicspath{{pictures/}}

\usetikzlibrary{calc}
\usepackage{zref-savepos}
\usepackage{tabu}

\aclfinalcopy 

\title{Topic Augmented Generator for Abstractive Summarization}

\author{Melissa Ailem$^{1}$,  Bowen Zhang$^{1}$ and Fei Sha$^{1,2}$   \\
$^{1}$ University of Southern California, Los Angeles, CA  \\
$^{1}$ {\tt  \{ailem, zhan734, feisha\}@usc.edu}\ \  $^{2}$ {\tt  fsha@google.com}
  }

\date{}

\begin{document}
\maketitle
\begin{abstract}
Steady progress has been made in abstractive summarization with attention-based sequence-to-sequence learning models.  In this paper, we propose a new decoder where the output summary is generated by conditioning on both the input text and the latent topics of the document. The latent topics, identified by a topic model such as  LDA, reveals more global semantic information that can be used to bias the decoder  to generate words. In particular, they enable the decoder to have access to additional word co-occurrence statistics captured at document corpus level. We empirically validate the advantage of the proposed approach on both the CNN/Daily Mail and the WikiHow datasets. Concretely, we attain strongly improved ROUGE scores when compared to state-of-the-art models.
\end{abstract}

\section{Introduction}
\label{intro}
Extractive summarization focuses on selecting parts (e.g., words, phrases, and sentences) from the input document~\citep{kupiec1999trainable, dorr2003hedge,nallapati2017summarunner}. While the primary goal is to preserve the important messages in the original text, the more challenging abstractive summarization aims to generate a summary via rephrasing and introducing new concepts/words~\citep{zeng2016efficient, nallapati2016abstractive,see2017get, paulus2017deep,liu2018generative}. All those entail a broader and deeper understanding of the document, its background story and other knowledge --- \emph{Zeitgeist} --- that are not explicitly specified in the input text but are nonetheless in the minds of the human readers.

Neural-based abstractive summarization has since made a lot of progress~\citep{nallapati2016abstractive, see2017get}. By large, the language generation component, i.e., the decoder, outputs summary by conditioning on the input text (and its representation through the encoder). 

\emph{What kind of information can we introduce so that richer texts can appear in the summaries}? In this paper, we describe how to combine topic modeling with models for abstractive summarization. Topics identified from topic modeling, such as Latent Dirichlet Allocation (LDA), capture corpus-level patterns of words co-occurrence and describe documents with mixtures of semantically coherent conceptual groups. Such usages of words and concepts provide valuable inductive bias for supervised models for language generation.

We propose Topic Augmented Generator (TAG) for abstractive summarization where the popular pointer-generator based decoder is supplied with latent topics of the input document~\cite{see2017get}. To generate a word, the generator learns to switch among conditioning on the text, copying the text, and conditioning on the latent topics. The latter provides a more global context to generate words. 

We apply the proposed approach on two benchmark datasets, namely CNN/DailyMail and WikiHow, and obtain strongly improved performance when the topics are introduced. Moreover, the summaries generated by our decoder have higher coherence with the original texts in the topic latent space, indicating a better preserving of what the input text is about.

\section{Approach}
\label{sApproach}
Our work builds on the popular attention-based neural summarization models. We review one such model~\cite{see2017get}, followed by the description of our approach.


\subsection{Attention-based Neural Summarization}
%
%

Let $\bx = (x_1,\ldots,x_{L})$ denote a document represented as a sequence of $L$ words. Similarly, let $\by = (y_1,\ldots,y_{T})$ denote a $T$-word summary of $\bx$. We desire $T\ll L$.

To learn the mapping from $\bx$ to $\by$ from a corpus of paired documents and their summaries, a natural formalism is to model the conditional distribution of $\by$ given $\bx$, i.e., $p(\by|\bx)$. Furthermore, generating the summary is Markovian, implying a factorized form of the distribution
\begin{equation}
p(\by|\bx) = p(y_1 | \bx) \prod_{t=2}^{T} p(y_t | \bx, y_{1:t-1})
\end{equation}
where $y_{1:t-1}$ stands for the generated $(t-1)$ words. Sequence-to-sequence (seq2seq) models typically use RNN encoder-decoder architectures to parameterize the above distribution.

The encoder reads $\bx$ word-by-word from the left to the right (and/or reversely with another encoder) and produces a sequence of hidden states $\{h_1, \ldots, h_i, \ldots, h_N\}$, with $h_i\in\mathbb{R^{K}}$ and $h_i = f_e(x_i, h_{i-1})$, where $f_e$ is a differentiable nonlinear function.
The decoder models the distribution of every word in the summary conditioned on the words preceding it and the encoder's hidden states
\begin{equation}
p(y_t|\bx, y_{1:t-1}) = \gamma_{t} = g_{y_t}(y_{t-1}, s_t, c_t)
\label{eDecoder1}
\end{equation}
where $g_{y_t}(\cdot)$ is a differentiable function (such as softmax over all possible words) that yields probability as output.  $s_t = f_d(y_{t-1}, s_{t-1}, c_t)$ is the decoder's current hidden state, with $f_d$ being a non-linear function. $c_t$ is the \emph{context vector}, a weighted average of the encoder hidden states:
\begin{equation}
c_t = \sum_{i=1}^{L}\alpha_{ti}h_i.
\end{equation}
The attention weights $\alpha_{ti}$, forming a categorical distribution $\balpha_t = (\alpha_{t1},\ldots,\alpha_{tL})$, capture the context dependency between input words $x_i$ and generated word $y_t$, cf.~\cite{bahdanau14}.  

\paragraph{Pointer-Generator (PG)} \citet{see2017get} modifies the generative probability eq.~(\ref{eDecoder1}) so that out-of-vocabulary words can be \emph{copied} from the input:
\begin{align}
p( y_t|\bx, y_{1:{t-1}})\! =\! p_t^{\textsc{pg}}\! =\! (1-\pi_t)\gamma_t + \pi_t\alpha_{t\,{y_t}} 
\label{ePG}
\end{align}
where the learnable switching term $\pi_t$  is adaptive during generation, and denotes the probability of using the attention distribution $\balpha_t$ to draw a word for the input document~\citep{see2017get}. 

\subsection{Topic Augmented Generator (TAG)}

\paragraph{Main idea} As discussed in the previous section, our main idea is to introduce bias into the decoder such that generating words is geared toward reflecting the broad, albeit latent, semantic information underlying the input document. 

With such bias, we hope the summary could include words that are ``exogenous'', but nonetheless semantically cohesive with the input document --- using such words is ``blessed'' by subscribing to explicitly learned word co-occurrence patterns at the document corpus level.

\paragraph{LDA}  To this end, we use Latent Dirichlet Allocation (LDA) model~\cite{blei03} to discover semantically coherent latent variables representing the input documents. 

The document is modeled as a sequence of sampling words from a mixture of $K$ topics, 
\begin{equation}
 p(\bx)\!=\!\!\int\!\!\prod_{i=1}^{L}\sum_{z_{x_i}}p(x_i|z_{x_i},\vbeta)p(z_{x_i}|\vtheta)p(\vtheta)d\vtheta    
\end{equation}
Given a corpus of text documents, the prior distribution for $\vtheta$ and the  topic-word vectors $\vbeta$ can be learned with maximum likelihood estimation. Additionally, for new documents, their (maximum a posterior) topic vector $\vtheta^*$ can also be inferred. The model parameter $\vbeta$ captures word co-occurrence patterns in different topics.

\paragraph{TAG}  We revise the conditional probability of eq.~(\ref{ePG}) with a new mixture component
\begin{align}
 p( y_t|\bx, y_{1:{t-1}})\!=\! 
 \lambda_t p_t^{\textsc{pg}} + (1-\lambda_t)q(\bmu_{y_t}^{\top}\vtheta^*)
\label{eTopicPG}
\end{align}
where $q(\cdot)$ denotes the softmax probability of generating $y_t$ according to the input document's topic vector $\vtheta^*$ and the topic-word distribution $\bmu_{y_t}$. Note that we can use without change the corresponding $\vbeta$ from the LDA model or use them as initialization and update them end-to-end. We adopt the latter option in our experiments.

The mixture weights $(1-\pi_t)\lambda_t, \pi_t\lambda_t$  and $(1-\lambda_t)$ are (re)parameterized with a feedforward neural net (NN) of a softmax output and learnt end-to-end too. 

\paragraph{Inference and learning} We use maximum likelihood estimation to learn the  model parameters. Other alternatives are possible~\cite{paulus2017deep} and left for future work.  

At  inference, we use LDA's parameters to infer topic vectors.  We do not include test samples in fitting LDA to avoid information leak.

\section{Related Work}
\label{sRelated}

Our work builds on the attention-based sequence-to-sequence learning models in the form of a pair of encoder and decoder~\cite{rush2015neural,chopra2016abstractive,nallapati2016abstractive}. Pointer-generator networks~\cite{see2017get,vinyals2015pointer,nallapati2016abstractive} were proposed to address out-of-vocabulary (OOV) words by learning to copy new words from the input documents. To  avoid repetitions, \citet{see2017get} used a coverage mechanism~\cite{tu2016modeling}, which discourages frequently attended words to be generated. Some work~\citep{liu2018generative,wang2018learning} explored adversarial learning to improve the quality of the generated summaries. \citet{wang2018learning} further consider a fully  unsupervised approach, which does not require access to document-summary training pairs.

\citet{wang2019topic} leverages topic information by injecting topic information into the attention mechanism. Specifically, they fix certain words's embeddings, derived from the LDA's topic-word parameters $\vbeta$.  However, they discard the document-level topic vector $\vtheta$, which plays a significant role in our model. Our early experiments with  injecting topics into attention mechanism yield only very minor improvement.




\section{Experimental Results}
\label{sExp}

\subsection{Setup}

\paragraph{Datasets}  We use two datasets
CNN/Daily Mail (CNN/DM) and WikiHow.

CNN/DM has been extensively used in recent studies on abstractive summarization~\citep{hermann2015teaching,see2017get, nallapati2017summarunner,wang2018learning}. It consists of 312,085 news articles, where each is associated with a multi-sentence summary (about 4 sentences).
As in~\cite{see2017get} we use the original version of this corpus, which is split into  287,227 instances for training, 11,490  and 13,368 for testing and validation respectively. 

WikiHow has been introduced recently as a more challenging dataset for abstractive summarization~\citep{koupaee2018wikihow}. It contains about 200,000 pairs of articles and summaries. The summaries are more abstractive  and are typically not the first few sentences of the documents. As in \citep{koupaee2018wikihow}, our  data splits consists of  168,128 ,  6000  and 6000 pairs for training, testing, and validation respectively. 


\paragraph{Methods in comparison}
Our main baseline is the seq2seq model with a pointer-generator~\citep{see2017get}, referred as {\bf Pointer-Generator (PG)} for short. We include another variant \textbf{PG+Cov} where the coverage mechanism is used to avoid repeating words receiving strong attentions~\citep{see2017get}.  Our approach has corresponding two variants: Topic Augmented Generator (\textbf{TAG}) and \textbf{TAG+Cov}. We also report the results of the {\bf Lead-3} baseline where the summary corresponds to the first three sentences of the input article. 

\paragraph{Evaluation metrics} We use mainly the ROUGE scores to evaluate the quality of generated summaries~\citep{lin2004rouge}. ROUGE-1, ROUGE-2 and ROUGE-L  measure respectively the  unigram-overlap,  bigram-overlap,  and the longest common sub-sequence between the predicted and reference summaries.

\paragraph{Model specifics} We follow the suggestion in ~\cite{see2017get}.  We use $K=100$ topics for both datasets. Details are in Supplementary Material.


\subsection{Quantitative results}

{\setlength\tabcolsep{7pt}
\def\arraystretch{1.05}
\begin{table*}[t]
    \caption{Comparison of various models on the test sets of CNN/DM and WikiHow. Higher scores are better.\\
     Models marked with (*) indicate results published in the original paper.}
    \label{cnn_res}
    \centering
    {\small
    \begin{tabular}{|c|c|c|c||c|c|c|}
      \cline{2-7}
  \multicolumn{1}{c}{}  &\multicolumn{3}{|c||}{CNN/Daily Mail} & \multicolumn{3}{c|}{WikiHow} \\
      \cline{2-7}
     \multicolumn{1}{c|}{}     & ROUGE-1  & ROUGE-2 & Rouge-L & ROUGE-1  & ROUGE-2 & Rouge-L\\
     \hline
      \textbf{PG*}~\cite{see2017get}    & 36.44  &15.66&33.42 & -& - &-\\
      \hline
      \textbf{PG+Cov*}~\cite{see2017get}   & 39.53&17.28&36.38  &  -&  -&-\\
            \hline
             \hline
   \textbf{PG} (by us)   & 35.73 & 15.08 & 32.69 &  26.02& 7.92  &24.59\\
        \hline
          \textbf{TAG} (this paper)    &  36.48&15.89&33.68  &   26.18 & 8.18 & 25.25 \\ 
        \hline
          \textbf{PG+Cov} (by us)     & 39.12  & 16.88  &  35.59 &27.08 &8.49& 26.25 \\ 
        \hline
          \textbf{TAG+Cov} (this paper)   &   \textbf{40.06} & \textbf{17.89}& \textbf{36.52}& \textbf{28.36} & \textbf{9.05} & \textbf{27.48} \\
        \hline
        \hline
                  \textbf{Lead-3}  & 40.34 &17.70 &36.57 &26.00 &7.24& 24.25   \\
     \hline
    \end{tabular}}
\end{table*}}

We report our main results in Table~\ref{cnn_res}. Clearly, models augmented with topics perform better than those who are not. 
More detailed analysis shows that  our model \textbf{TAG+Cov} performs better that \textbf{PG+Cov} on more than half of the test documents (5888 out of 11490 on CNN/DM). We give details in the Supplementary Material.

\begin{figure}[t!]
   \centering
    \includegraphics[height = 4.5cm, width = 7cm]{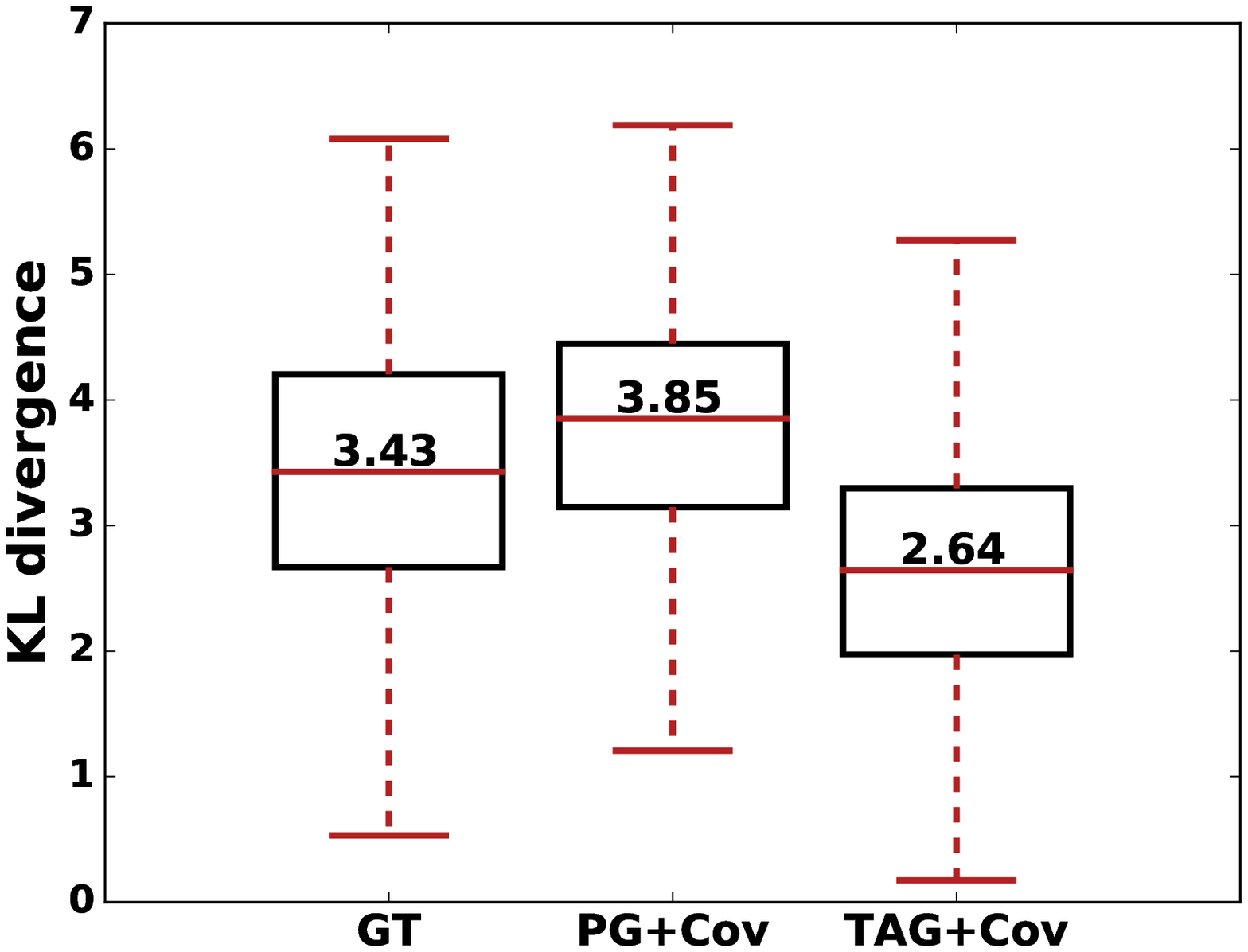}
     \includegraphics[height = 4.5cm, width = 7cm]{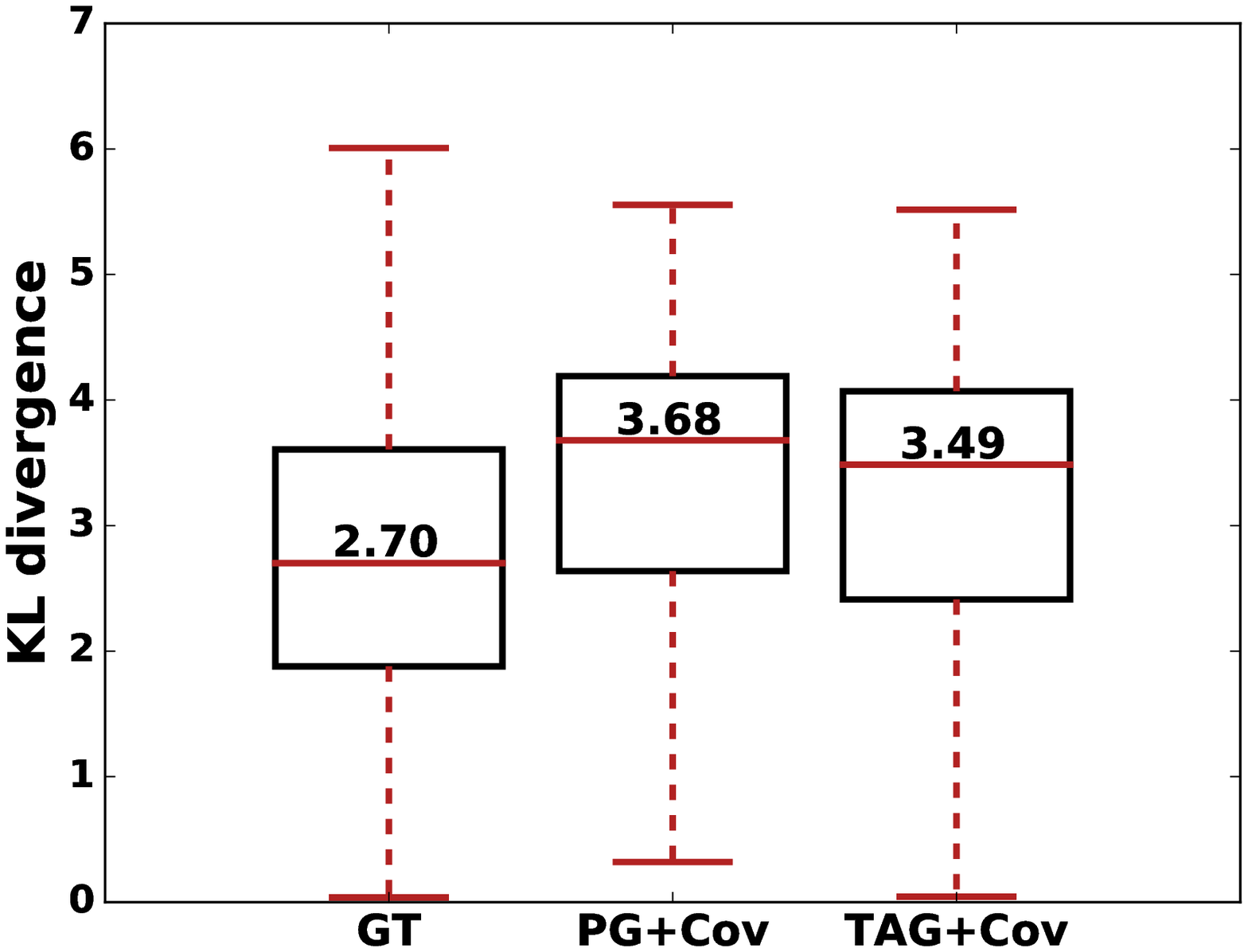}
 \caption{\small Top (CNN/DM), Bottom (WikiHow). Boxplots of pairwise KL divergences in the topic distributions $\btheta^*$ between the original documents and: the Ground Truth (GT), generated summaries. Lower implies better semantic coherence.}

    \label{cnndmtopics_}
\end{figure}

A good summary should also preserve well the main topics in the original document. To assess this aspect, 
for each test document $\bx$, we compute the Kullback-Leibler divergence between the topic distributions $\btheta^*$ inferred on the original document and on the reference as well as the generated summaries. As shown by the boxplots (\ie, median, minimum, maximum, first and the third quantiles) in Figure~\ref{cnndmtopics_}, \textbf{TAG+Cov} tends to generate summaries that are more coherent with those of the original documents. In the case of CNN/DM, what is particularly interesting is that the summaries by the \textbf{TAG+Cov} turn out to be more semantically coherent with the input texts than the ground-truth summaries. We suspect that the (ground-truth) summaries for news articles are likely too concise for a topic model to detect reliably co-occurrence patterns needed for inferring topics.
\subsection{Qualitative results}
In Figures~\ref{ex2} and \ref{ex4}, we present two examples. In the first one, our model generates 3 main sentences, instead of 2 by the other model. Both have missed the last important sentence in the ground-truth summary.  In the second example, \textbf{PG+Cov} misses a lot important words. More examples are given in the Supplementary Material. Overall, it seems like our model is more likely to capture the main topics of the initial documents and tend to be more concise.
\begin{figure}[t!]
    \centering
    \begin{tabular}{|p{3.1in}|}
    \hline
    {\small
       \textbf {Ground Truth:}  british physicist stephen hawking has sung monty python 's galaxy song . song is being released digitally and on vinyl for record store day 2015 .
 it is a cover of the song from 1983 film monty python 's meaning of life . professor hawking , 73 , appeared on film alongside professor brian cox . }\\
        \hline
       {\small \textbf {PG+Cov:} british physicist stephen hawking has sung monty python 's galaxy song -lrb- clip from the video shown -rrb- . one of the world 's greatest scientists has covered monty python 's classic galaxy song , 
taking listeners on a journey out of the milky way . it is a cover of the song from 1983 film monty python 's meaning of life .}\\
        \hline
           {\small\textbf {TAG+Cov:} british physicist stephen hawking has sung monty python 's galaxy song -lrb- clip from the video shown -rrb- . the song is being released digitally and on vinyl for record store day 2015 .
it is a cover of the song from 1983 film monty python 's meaning of life .  }\\   
          \hline
    \end{tabular}
    \caption{\small From CNN/DM.  Our TAG+Cov model generates 3 main sentences instead of 2 for PG+Cov.}
    \label{ex2}
\end{figure}

\begin{figure}[t!]
    \centering
    \begin{tabular}{|p{3.1in}|}
    \hline
    {\small
       \textbf {Ground Truth:} acquire a pot. gather the ingredients needed to make the curry. walk to the pot on your kitchen counter. choose the ingredients that you need for the recipe.}
   \\
        \hline
        {\small \textbf {PG+Cov:} go to the kitchen counter. go to your kitchen counter. look for the ingredients you want to cook. finished. take care of your health.}\\
        \hline
           {\small \textbf {TAG+Cov:} acquire a pot. gather the ingredients needed to make the dish. walk to the pot on your kitchen counter. choose the ingredients that you need to add to your recipe. confirm your decision.}\\
        \hline
    \end{tabular}
    \caption{\small From WikiHow (``How to Make Vegetable Curry in Harvest Moon Animal Parade'').}
    \label{ex4}
\end{figure}

\section{Conclusion}
\label{sDiss}
We have shown that by conditioning on the topics underlying the input documents, the decoder generates noticeably improved summaries. This suggests that the supervised learning of representation from sequence-to-sequence models can benefit from unsupervised learning of latent variable models. We believe this is likely a fruitful direction for the task of abstractive summarization where rephrasing and introducing new concepts that are not observed in the input texts are essential.

\bibliography{emnlp-ijcnlp-2019}
\bibliographystyle{acl_natbib}

\end{document}